\documentclass[letterpaper, 10 pt, conference]{ieeeconf}  % Comment this line out if you need a4paper

\IEEEoverridecommandlockouts                              % This command is only needed if 
\usepackage{graphicx}
\usepackage{subfigure}
\usepackage{hyperref}
\usepackage{url} 
\usepackage{xcolor}
\usepackage{setspace}
\usepackage[ruled,linesnumbered,vlined,norelsize,\languagename]{algorithm2e}
\SetAlFnt{\footnotesize}
\SetAlCapFnt{\small}
\SetAlCapNameFnt{\small}

\title{\LARGE \bf
Learning the Sequence of Packing Irregular Objects from Human Demonstrations: 
Towards Autonomous Packing Robots}
\author{André Santos$^{1}$, Nuno Ferreira Duarte$^{1}$, Atabak Dehban$^{1}$ and José Santos-Victor$^{1}$% <-this % stops a space
\thanks{*This work was supported by the Fundação para a Ciência e a Tecnologia (FCT) through the  ISR/LARSyS Associated Laboratory UID/EEA/50009/2020, LA/P/0083/2020.}% <-this % stops a space
\thanks{$^{1}$All the authors are affiliated with the Institute for Systems and Robotics, Instituto Superior Técnico, Universidade de Lisboa, 1049-001 Lisbon, Portugal (e-mail: andrejfsantos@tecnico.ulisboa.pt, \{nferreiraduarte, adehban, jasv\}@isr.tecnico.ulisboa.pt).}%
}

\begin{document}

\maketitle
\thispagestyle{empty}
\pagestyle{empty}

%%%%%%%%%%%%%%%%%%%%%%%%%%%%%%%%%%%%%%%%%%%%%%%%%%%%%%%%%%%%%%%%%%%%%%%%%%%%%%%%
\begin{abstract}

We tackle the challenge of robotic bin packing with irregular objects, such as groceries. Given the diverse physical attributes of these objects and the complex constraints governing their placement and manipulation, employing preprogrammed strategies becomes unfeasible.
Our approach is to learn directly from expert demonstrations in order to extract implicit task knowledge and strategies to ensure safe object positioning, efficient use of space, and the generation of human-like behaviors that enhance human-robot trust.
We rely on human demonstrations to learn a Markov chain for predicting the object packing sequence for a given set of items and then compare it with human performance. Our experimental results show that the model outperforms human performance by generating sequence predictions that humans classify as human-like more frequently than human-generated sequences. 
The human demonstrations were collected using our proposed VR platform, ``BoxED'', which is a box packaging environment for simulating real-world objects and scenarios for fast and streamlined data collection with the purpose of teaching robots. We collected data from 43 participants packing a total of 263 boxes with supermarket-like objects, yielding 4644 object manipulations. Our VR platform can be easily adapted to new scenarios and objects, and is publicly available, alongside our dataset, at
\url{https://github.com/andrejfsantos4/BoxED}.
\end{abstract}

%%%%%%%%%%%%%%%%%%%%%%%%%%%%%%%%%%%%%%%%%%%%%%%%%%%%%%%%%%%%%%%%%%%%%%%%%%%%%%%%

\section{INTRODUCTION}

The ceaseless digitalization of the modern world has steadily transformed many aspects of our daily lives, including the grocery shopping experience.
In recent years, we have observed a pronounced migration from physical retail spaces to the online realm, with companies such as Ocado~\cite{Ocado}, a dedicated online grocery retailer, reporting 2.5 billion GBP of revenue in 2022.
This trend is underpinned by the intricate logistical challenge of efficiently packaging all orders into shipping containers.

\begin{figure}[t]
    \centering
    \includegraphics[width=0.35\textwidth]{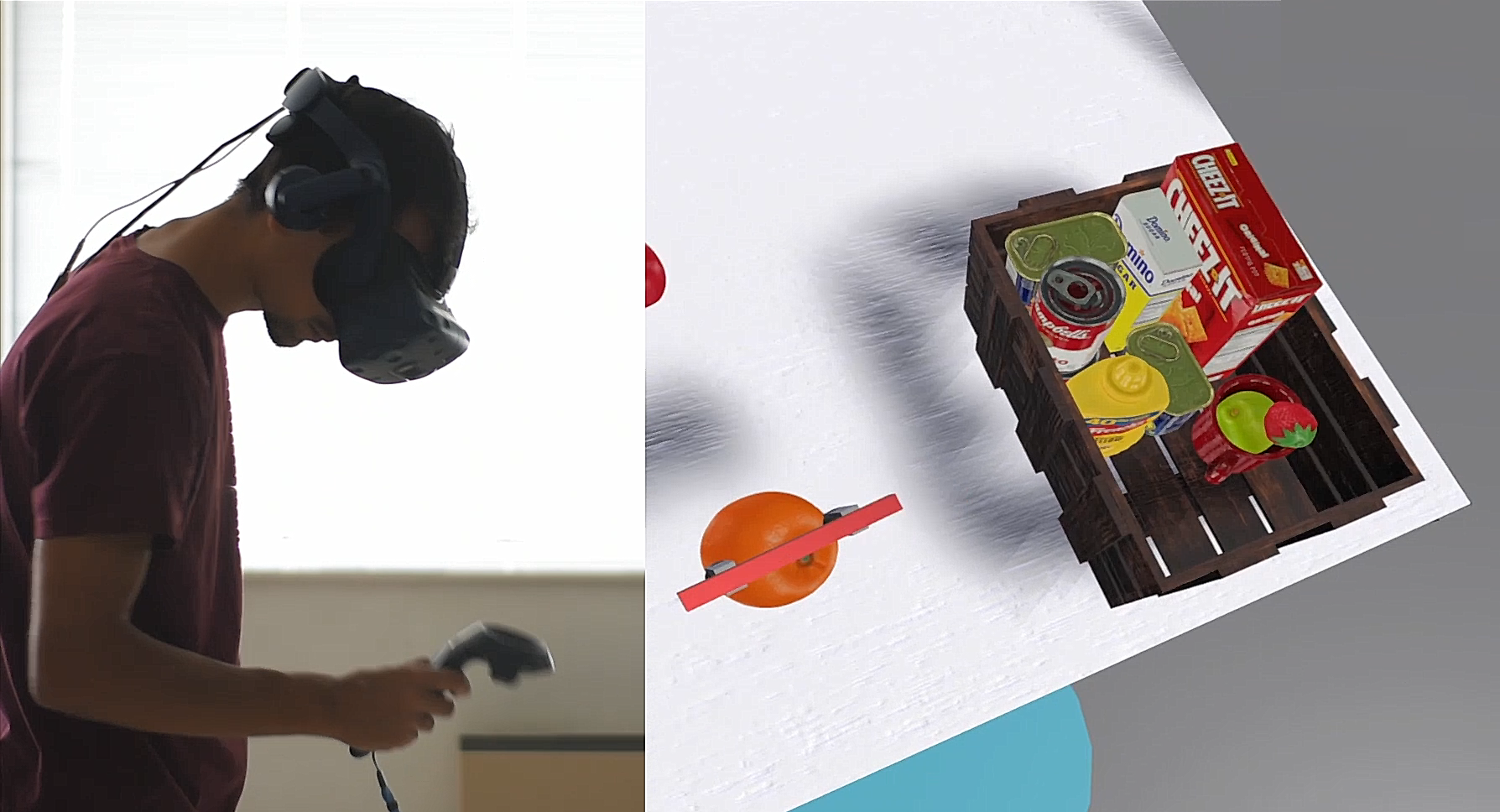}    
    \caption{The demonstrator performs the box packing task in virtual reality.}
    \label{fig:cover_img}
\end{figure}

Simultaneously, robots have been steadily integrating into both our workplaces and homes. The so-called ``cobots''~\cite{Cobots} have attracted a considerable amount of research and development attention, with the introduction of Amazon's new Astro robot~\cite{AmazonAstro} and Temi robot~\cite{TemiRobot}.
Besides the critical need to ensure a safe operation while interacting with humans, robots must also act as human-like as possible, such that their actions can be understood and anticipated by their interaction partners, providing users with an improved interaction experience.

The task that motivated our work lies at the intersection of these two rapidly expanding fields: a robot that packs groceries alongside and learns from a human collaborator. 
Several hardware solutions have been developed specifically to enhance robot-human collaboration, featuring special adaptations such as flexible joints. %Furthermore, the grasp inference literature contains many algorithms with impressive accuracy \cite{SurveyGrasping2020}. 
However, there is very little literature on how to pack heterogeneous objects such as those found in a supermarket, and there is even less publicly available datasets related to this task.
Existing methods that address packing objects inside a container consider essentially simple shape objects (such as cuboids and spheres) and often deploy search-based algorithms to determine the best placement pose, requiring multiple hours to find a solution \cite{RobotPackingKnownItems, StableBinPacking}.
Methods that predict a packing sequence given a set of objects (for instance, the objects in an online order) are equally sparse \cite{gyorgy2010line}.
Our contributions are twofold:
\begin{enumerate}
    \item A model trained on our dataset that predicts the correct packing sequence, extracting implicit task knowledge directly from humans. The generated sequences lead to safe and efficient packing and were classified as ``human-like'' sequences by users. %The generated sequences lead to safe and efficient positioning of the objects and are classified by experts as human-like sequences.
    
    \item A publicly-available VR platform (BoxED) for fast and streamlined data collection, see Fig.~\ref{fig:cover_img}. BoxED can be adapted to new box packaging scenarios with any object datasets, it collects 6-DOF pick-and-place grasp poses, object trajectories, packing sequences, and objects' poses when inside the box. In addition to BoxED, we created the first publicly available collection of human demonstrations of packing groceries into a box.
    
\end{enumerate}

Section~\ref{related_work} addresses related research, Section~\ref{dataset_collection} describes the dataset, Section~\ref{sequence_prediction} introduces our packing sequence prediction model, and Section~\ref{conclusions} presents final remarks.

\section{Related Work}
\label{related_work}

\noindent \textbf{Learning from Demonstrations (LfD).} LfD consists of learning how to execute new tasks and their constraints by observing an expert performing them \cite{LfDsurvey}.
This paradigm has attracted increasing attention from the robotics community because it circumvents predefined behaviors and introduces more flexibility in robotic learning.
These methods require tracking the human's actions which is not a trivial task due to rapid movements, occlusions, and observation noise~\cite{EndtoEndDrivingViaImitation2018,BayesianDisturbanceInjection2021}.
% 
%Recently, policy learning using deep learning methods have been used to extract knowledge from demonstrations ~\cite{VisionBasedMultiTaskManipulation2018,LearningTaskConstraints2021}.  
%One common approach for extracting knowledge from demonstrations is policy learning, where the goal is to learn a policy that maps the input space into the output space. %~(usually the state and action spaces, respectively) and that can accurately mimic the demonstrations.
%These approaches bypass explicit state characterization by defining as input raw data from the system, while others such 
%Another approach that is widely used is based on Markov Decision Processes.
%For instance, 
Munzer~T.~et~al.~\cite{EfficientBehaviorLearning2018} used a variation of Markov Decision Processes along with first-order logic in a system for a collaborative robot that learns tasks and human preferences before and during execution.
A major challenge in LfD is how to encode the demonstration in a useful format for training models.
In some works, participants wear a motion capture suit \cite{HumanCenteredCollaborativeRobots2021}, or the demonstrations occur in simulation \cite{TransformersOneShotVisImit2020}, while others use complex pose estimators \cite{DemoGraspFewShotLearning2021}. 

\noindent \textbf{Bin Packing.} It addresses the task of packing a set of objects into a bin with maximum space efficiency.
In general, there are two categories of methods: those based on search strategies and those that use unsupervised learning methods such as reinforcement learning (RL).
The first cluster offers the advantage of rapid deployment due to the absence of a learning phase, but the majority of cases suffer from slow execution times (ranging from dozens of seconds to a few hours \cite{RobotPackingKnownItems, StableBinPacking}). 
Conversely, methods based on RL or Deep RL (DRL)~\cite{BinPackingConfigTrees, GeneralizedRLBinPacking} require complex training schemes but have faster execution times.
Our approach aims to overcome a strong limitation in both categories of methods, as they often consider solely cuboid objects (otherwise the search and action spaces would grow drastically) and ignore properties such as fragility.  

\noindent \textbf{Sequence Prediction.} 
Various types of sequence prediction techniques exist, including sequence-to-sequence and sequence classification. However, our specific emphasis lies in predicting the subsequent element within a sequence, which encompasses tasks like time series forecasting and product recommendation.
Algorithms designed to tackle this category of problems exhibit a broad spectrum of approaches, from explicit association rules~\cite{Apriori} and pattern mining algorithms~\cite{Spade} to deep-learning based approaches that learn implicit representations. These algorithms search for sub-sequences that appear often in the data and thus contain important sequential or associative information to predict the next element.

\noindent \textbf{Related Datasets.} Perhaps most similar to our work is the dataset proposed by Song S. et al.~\cite{GraspingWildLearning2020}, consisting of videos of eight participants performing pick-and-place tasks in cluttered environments. The videos are annotated with the gripper trajectory (before the grasp and during the object manipulation), grasp pose, picking order, and object mask. Although more encompassing than most robotic grasping datasets, this dataset is not related to bin packing and as such does not include packing order, placement inside a container or object pose since pose estimation in a real environment is not trivial.

%%%%%%%%%%%%%%%%%%%%%%%%%%%%%%%%%%%%%%%%%%%%%%%%%%%%%%%%%%%%%%%%%%%%%%%%%%%%%%%%
\section{The BoxED Platform}
\label{dataset_collection}

\subsection{The Virtual Environment}

\begin{figure}[thpb]
    \centering
    \includegraphics[width=0.35\textwidth]{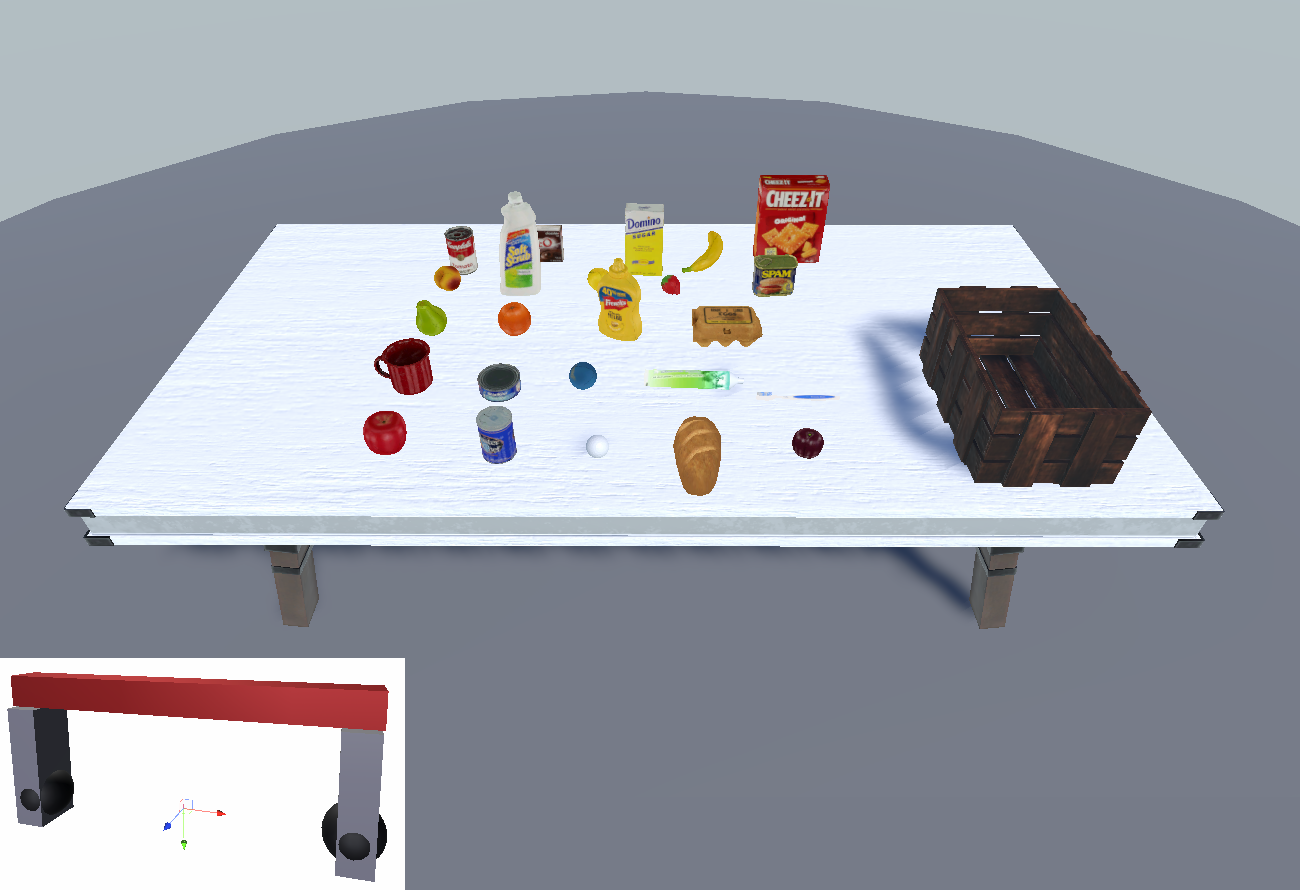}    
    \caption{The environment and robot gripper created for packing groceries.}
    \label{fig:unity_env}
\end{figure}

A virtual reality (VR) environment was created in Unity~\cite{Unity}, illustrated in Figure~\ref{fig:unity_env}, that consists of a circular area where the user is free to move and a table in its center. The task takes place on the table: the user should pack the available objects into a box\footnote{The supplementary material presents a participant completing the task.}.
We use 24 different objects, most chosen from the YCB dataset~\cite{YCB_dataset} and some obtained from public object model platforms.

The participants interact with the virtual world via the physical controller.
The pose of the controller is mimicked by the virtual gripper and a pressure-sensing button controls the closing and opening of the virtual gripper's fingers.
The objects are configured as rigid bodies, which means that they are affected by gravity and friction, and can collide with other objects.
Furthermore, we implemented haptic feedback on the physical controller to signal collisions between the gripper and objects to the participants.
Upon grasping an object, it is rigidly attached to the gripper so that its relative pose is maintained throughout the manipulation. 

\subsection{Data Collecting Experiments}

\begin{figure}[thpb]
    \centering
    \includegraphics[width=0.30\textwidth]{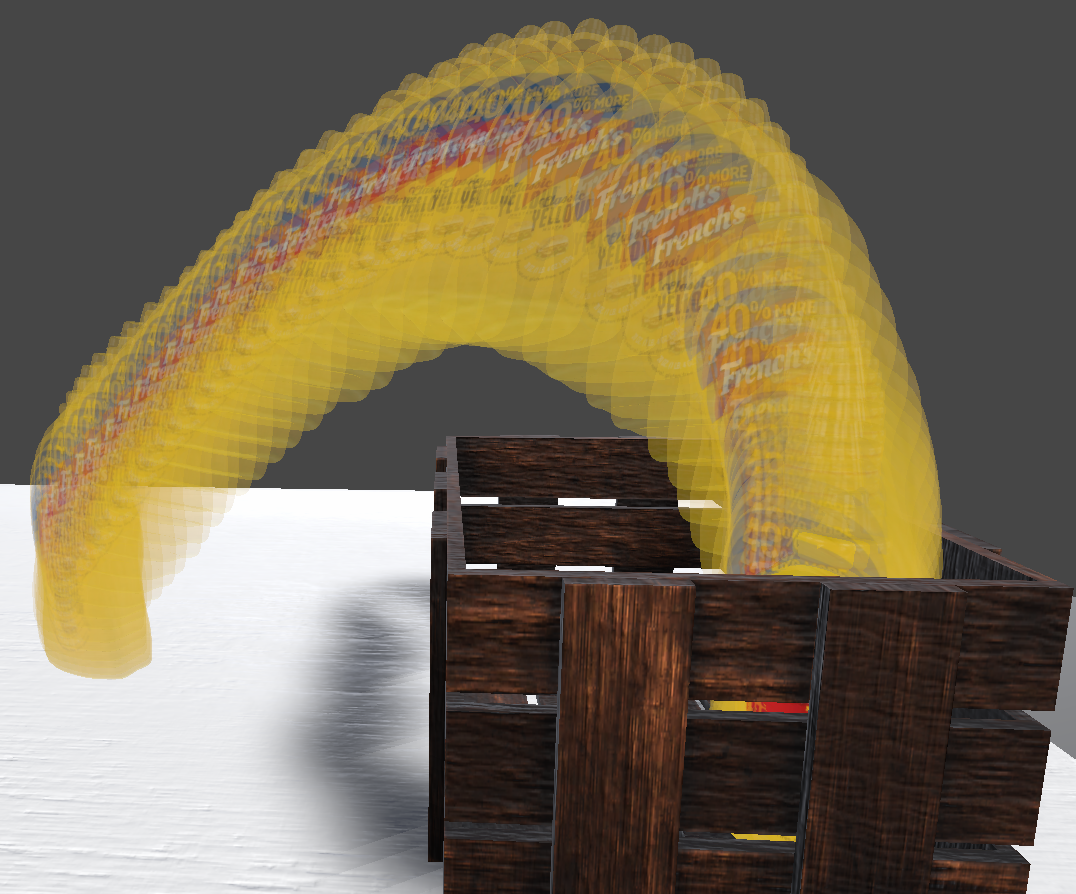}    
    \caption{An example object trajectory from BoxED. One semi-transparent copy of the object (a mustard bottle) was spawned at each recorded pose.}
    \label{fig:obj_traj}
\end{figure}

We chose to collect the BoxED dataset in VR as it simplifies much of the complexity associated with tracking objects and gripper pose seen on \cite{GraspingWildLearning2020}.
Before starting the task (i.e., the scene), each participant was instructed to pack all the items into the box in a manner similar to how they would typically pack groceries in a supermarket.
%Particular emphasis was given to the objects' fragilities since this property is not present in the virtual environment but must be considered when stacking objects on top of each other.
They were also instructed to pack the objects in an orderly manner starting from one side of the container, for reasons that are clarified in Section~\ref{section:markov_model}.
It is important to mention that the objects presented here constitute a random subset of the complete object collection. These objects were generated in a way that ensures the total combined volume of each object's bounding box falls within the range of 70\% to 90\% of the container's volume.
This is to ensure that the task is not trivial, forcing the participants to carefully reflect on how to place the objects. The initial poses of the objects are spread out as shown in Figure \ref{fig:unity_env}.
%At the beginning of the scene the objects are spread out on the table with no overlaps to ensure that the participant can see and grasp all objects.
%
The experiment consists of 4 different scenes to pack a box. 
For each object manipulation, we record the 6-DOF pick-up grasp pose, the 6-DOF pose inside the box, and the object's trajectory from the former to the latter sampled at 20 Hz and shown in Fig.~\ref{fig:obj_traj}.
Additionally, we also save the sequence in which the objects were packed, the pose of the VR headset throughout the experiment, and a top-down view of the initial layout of the objects.

\subsection{Unpacking the Collected Data}

\begin{figure}[htbp]
\centering
    \subfigure
    {
        \label{fig:histogram_dur}
        \includegraphics[width=0.22\textwidth]{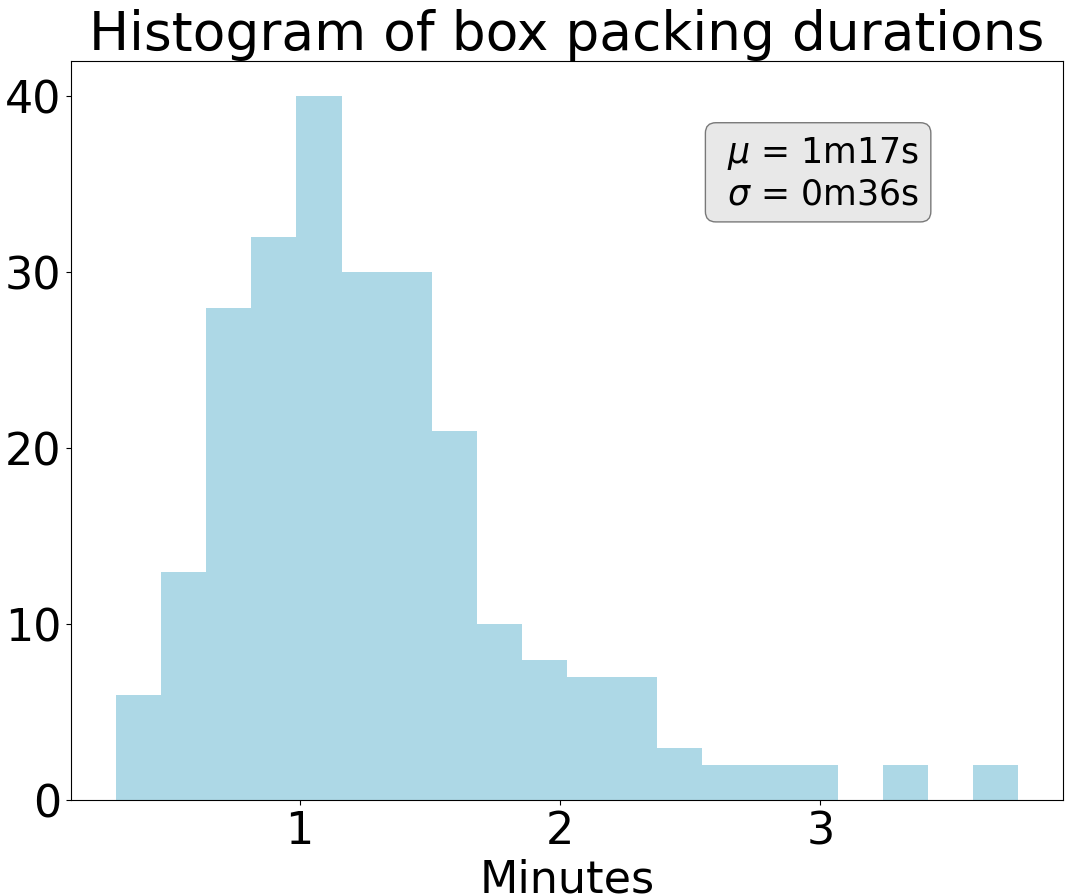}
    }  
    \centering 
    \hfill
    \subfigure
    {
        \label{fig:dur_vs_objs}
        \includegraphics[width=0.228\textwidth]{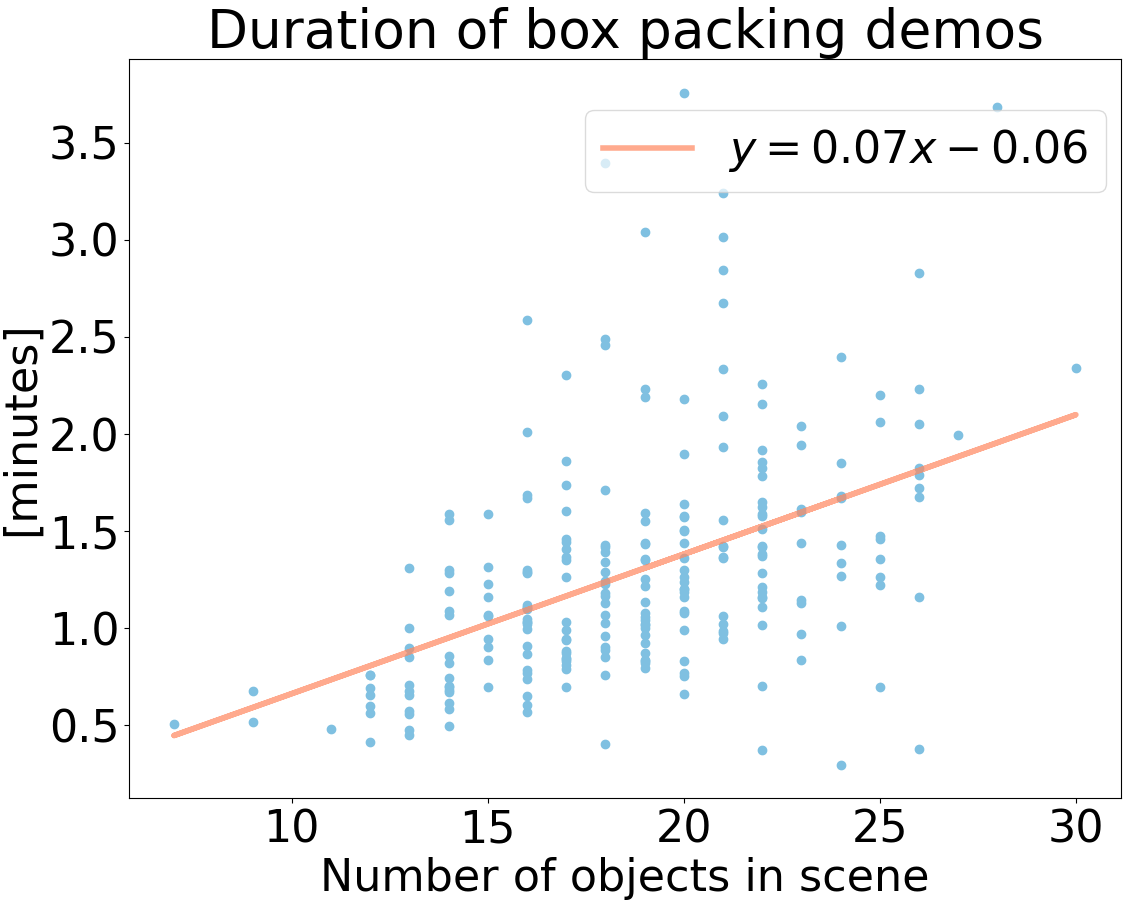}
    }   
    \caption{(Left) Histogram of all scenes' durations. (Right) Duration of each scene as a function of the number of objects. The pink line was obtained with a linear regression.}
\end{figure}

The experiments were conducted using an HTC Vive Pro headset, totaling 75 experiments and 43 different participants~(some participants volunteered more than once).
In total the dataset contains 263 scenes, translating into 4644 grasp poses and object placements, which on average corresponds to 194 grasp poses and placements per object. In the histogram of scene's duration, presented in Fig.~\ref{fig:histogram_dur}, shows that it approximates a gamma distribution with an average duration of 1 minute and 17 seconds, and a standard deviation of 36 seconds. 
Furthermore, when plotting scene's duration as a function of the number of objects it indicates that, on average, each object adds 4 seconds to the task's duration. 
This is expected since the difficulty of the task naturally increases with the number of objects to pack.

\begin{figure}[htbp]
\centering
    \subfigure[]
    {
        \label{fig:3objs}
        \includegraphics[width=0.08\textwidth]{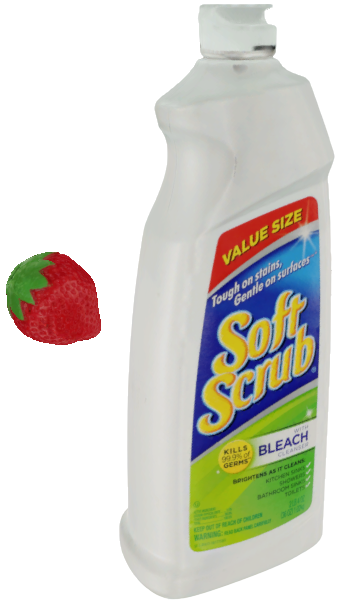}
    }  
    \subfigure[]
    {
        \label{fig:place_pos_strawberry}
        \includegraphics[width=0.176\textwidth]{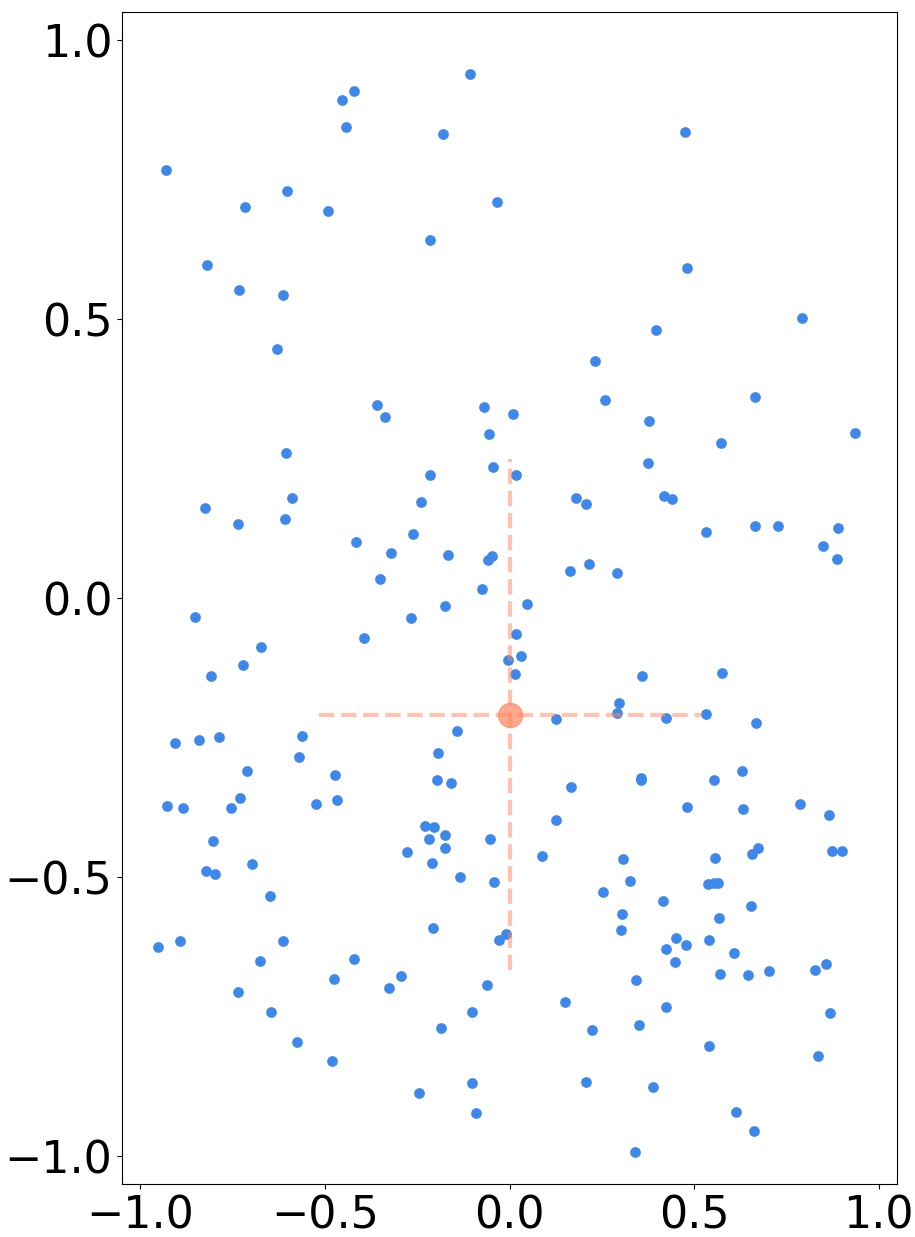}
    }  
    \centering  
    % \hfill
    \subfigure[]
    {
        \label{fig:place_pos_bleach}
        \includegraphics[width=0.176\textwidth]{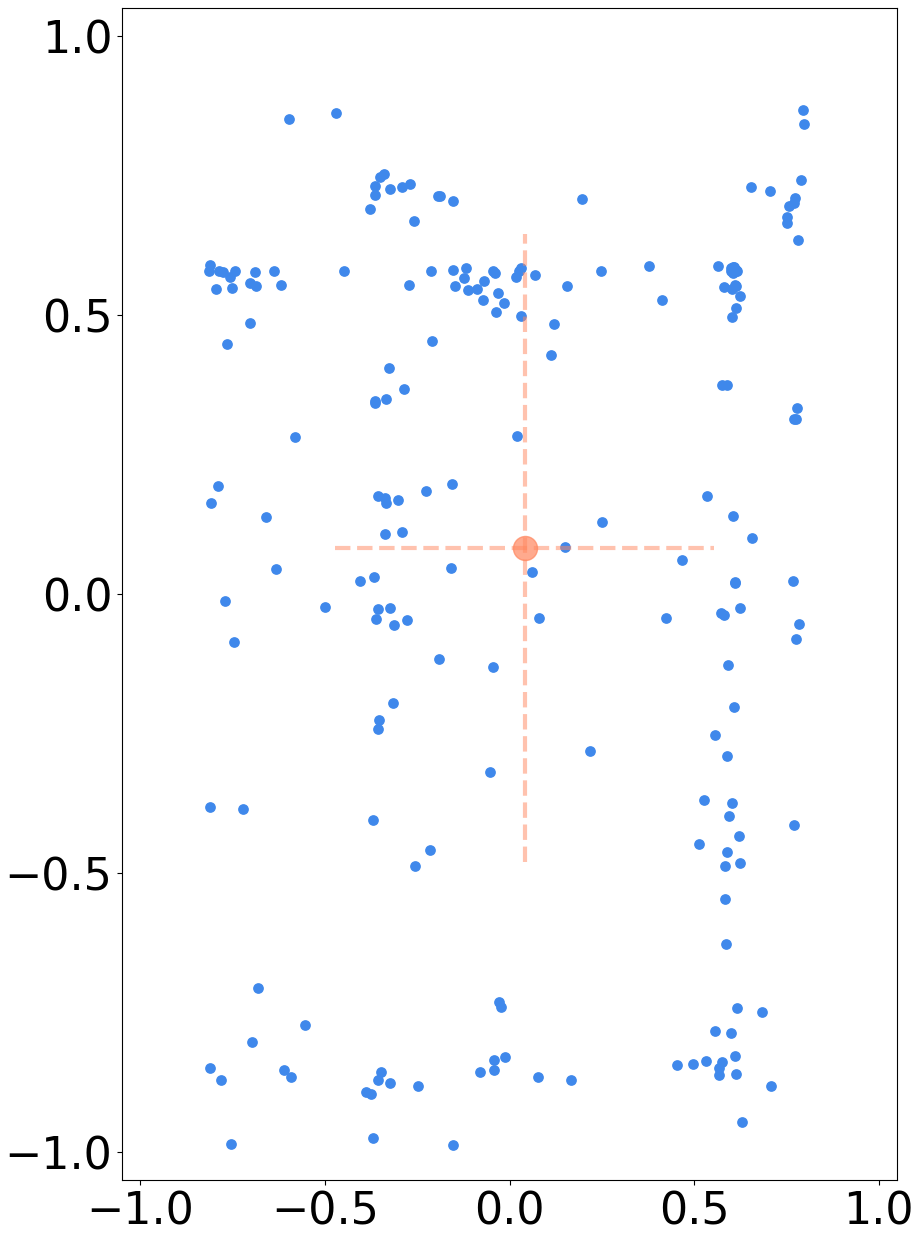}
    } 
    \caption{(a) Objects considered in placement pose analysis: strawberry and bleach cleanser. (b-c) Corresponding placement positions inside the box of these two objects, from a top-down view with normalized coordinates. The larger pink circle is the average position and each dashed line is the standard deviation along the corresponding axis.}
    \label{fig:place_pos}
\end{figure}

Analyzing the placement poses is not trivial since they are six-dimensional variables.
Thus, for better visualization, we analyze the position and orientation separately, and only the two objects shown in Fig.~\ref{fig:3objs}. 

% \begin{figure}[htbp]
% \centering
%     \subfigure[]
%     {
%         \label{fig:strawberry}
%         \includegraphics[width=0.03\textwidth]{./Images/strawberry_}
%     }    
%     \hspace{1cm}
%     \subfigure[]
%     {
%         \label{fig:bleach_cleanser}
%         \includegraphics[width=0.051\textwidth]{./Images/bleach_cleanser}
%     } 
%     \caption{Objects considered in placement pose analysis: (a) strawberry and (b) bleach cleanser.}
%     \label{fig:3objs}
% \end{figure}

To detect frequent patterns in the positioning of different objects inside the box, Fig.~\ref{fig:place_pos} shows a plot of the placement positions for two different objects, in a top-down view.
In other words, this plot ignores the height of the placement and considers only the horizontal coordinates inside the box.
The strawberry is placed uniformly throughout the box, however the average placement is closer to the side of the box closest to the participant.
This occurs because the strawberry is one of the most fragile items and so participants tended to pack it only after all other objects, confirming that participants considered the objects' fragilities during the task.
In contrast, the bleach cleaner has a very uneven distribution.
It is clear that it is generally placed along the edges of the box and further away from the participant.
This is likely a strategy that most participants intuitively adopted due to two factors.
Firstly, to optimize the available space inside the box, and secondly because large objects would become an obstacle for the task if they were placed closer to the participant since they would obstruct subsequent placements.

\begin{figure}[htbp]
\centering
    \subfigure
    {
        \label{fig:place_rot_strawberry}
        \includegraphics[width=0.16\textwidth]{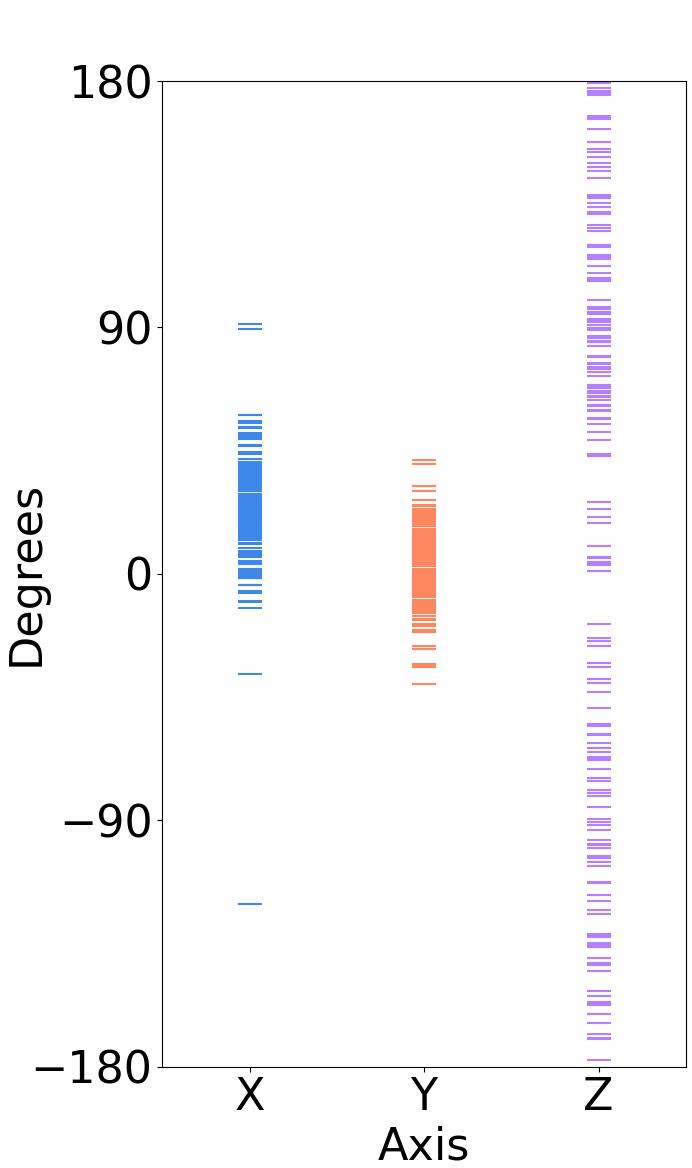}
    }  
    \centering 
    % \hfill
    \hspace{0.01\textwidth}
    \subfigure
    {
        \label{fig:place_rot_bleach}
        \includegraphics[width=0.16\textwidth]{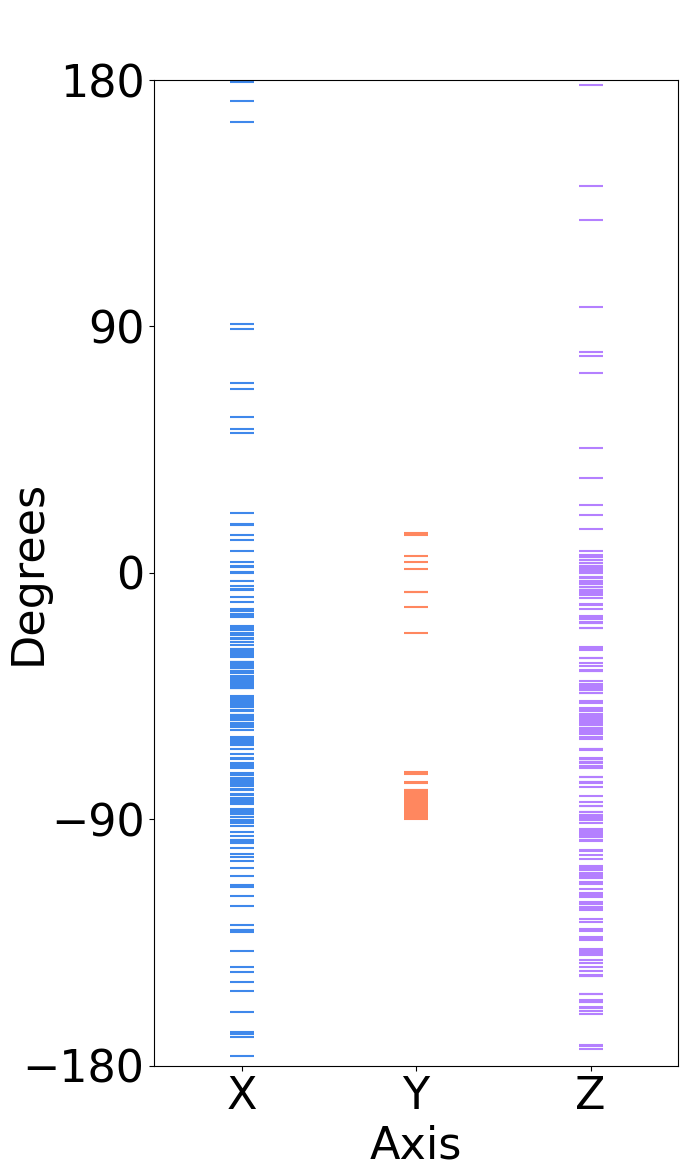}
    } 
    \caption{Placement orientations inside the box of two objects, where each line segment corresponds to one placement. The objects are strawberry (left) and bleach cleanser (right).}
    \label{fig:place_rot}
\end{figure}

A similar strategy can be used to analyze the orientation, except in this case, we visualize independently each of the three angles that compose it - rotation around each axis. This analysis is presented in Fig.~\ref{fig:place_rot}.
The visualization presented in Fig.~\ref{fig:place_rot} reveals, once again, distinctive patterns for each object.
For instance, both have near uniform distribution around the Z axis since they can be safely placed in any orientation around it.
On the other hand, the bleach cleanser presents a very skewed distribution around the Y axis~(which in Unity points up).
This is a consequence of its symmetric shape which leads participants to frequently place it aligned with the edges of the box.

The above two examples reemphasize the vast implicit knowledge humans use to pack a box with groceries and motivate our approach of learning these patterns from the data rather than relying on hand-designed rules or constraints.

%%%%%%%%%%%%%%%%%%%%%%%%%%%%%%%%%%%%%%%%%%%%%%%%%%%%%%%%%%%%%%%%%%%%%%%%%%%%%%%%
\section{Packing Sequence Prediction}
\label{sequence_prediction}

In this section we present both the approach we used to learn the packing sequence model from human demonstrations, as well as how to use the model to predict packing sequences for a certain set of objects.

\subsection{Modeling Choice and Assumptions}

Markov chains are popular because of their simplicity and by not requiring large amounts of data to train. 
This latter fact is important since our dataset contains less than 300 object-packing sequences, which is not ideal for deep learning. Generally in these approaches, the state encodes relevant information about the target environment and agent at each instant, and the agent's actions cause state transitions. 
To deploy such a model to predict the packing sequence we consider the following simplifying assumptions:
\begin{enumerate}
    \item The choice of what object to pack next depends only on the last object.
    \item The aforementioned choice does not depend on the pose of the last object.
\end{enumerate}

Furthermore, we can ignore the restrictions imposed by what objects are still available to pack for now~(we show later how we address this when sampling the Markov chain).
Consequently, the state is defined as the object's name and transitioning to a new state corresponds to packing another object.
This may be considered an over-simplification of our problem given that, in fact, when deciding what object to pack next, a human will likely take into account several factors such as: the last few objects that were packed~(not just the last one), their disposition inside the box, and the objects that remain unpacked. Nonetheless, we show in this section that our simplified formulation can achieve high accuracy.

\subsection{Learning the Packing Sequence Prediction Model}
\label{section:markov_model}
To formalize the model, let $O=\{obj_A, obj_B, ...\}$ be the set of all objects. We define $S=\{{<}start{>}, O\}$ as the set of all possible states, where at each instant the state indicates what object was packed last, and the state ${<}start{>}$ indicates that no object was packed inside the box yet.
Each transition probability, from an object $obj_A$ to $obj_B$, is represented by the conditional $P(obj_B\:|\:obj_A)$ and indicates the likelihood of placing $obj_B$ after placing $obj_A$.
To materialize an actual model, we need to estimate the transition probabilities.
Recall from the previous section that participants were asked to pack the box orderly, such that if an object ``A'' is placed after object ``B'', then ``A'' is likely stacked on top of ``B'', or at least is next to it.
This request enables the extraction of meaningful patterns from the sequences without having to analyze each object's placement pose. 
%
%However this is not true for all pairs of objects in the sequences, and so we need to extract only the meaningful patterns in the dataset.
As we assumed that choosing the next object depends only on the last, we we will only extract ordered pairs with the format $obj_A\rightarrow obj_B$ from the sequences. 
Inspired by pattern mining algorithms~\cite{Spade}, we extracted the pairs of objects that appeared in the sequences with a frequency above an adjustable threshold (0.032).
In other words, the pairs of objects whose support (probability of appearing in a sequence in the dataset) is smaller than the threshold are discarded. %The adjustable threshold controls how sensitive the model is to the transitions in the data.
A higher threshold retains only the most frequent transitions whereas a lower threshold retains more transitions and produces a Markov chain that is more densely connected, at the cost of including transitions that may be irrelevant. 
This threshold was adjusted such that it was small enough to include at least one transition for each state in $S$ and large enough to minimize loops in the chain, in order to maintain a hierarchical structure.

\begin{figure}[htbp]
    \centering
    \includegraphics[width = 0.80\linewidth]{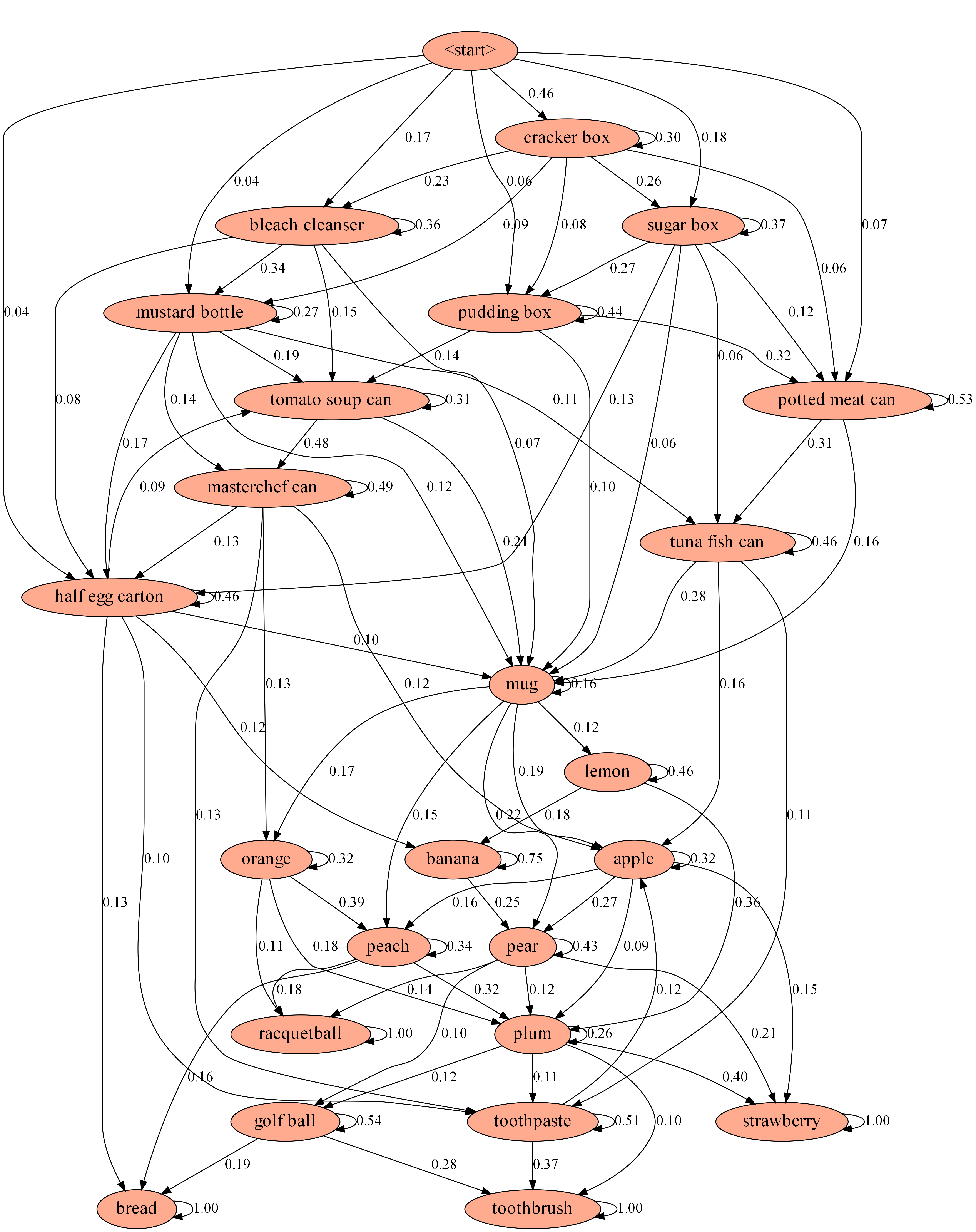}
    \caption{A Markov chain modeling the object packing sequence. Each node corresponds to a state and the numbers on the transition arrows are the corresponding probabilities.}
    \label{fig:markov_chain}
\end{figure}

Once the transition probabilities were normalized such that the sum of the transition probabilities from each state is 1, the resulting Markov chain is presented in Fig.~\ref{fig:markov_chain}.
The Markov chain placed first larger and more robust objects, followed by smaller and more fragile objects.
For instance, the topmost object is the cracker box, which is the largest object in the dataset and was usually placed first inside the box by the participants.
On the other hand, the bread and the toothbrush are the bottom-most objects since they are the most fragile and smallest, respectively.
%Furthermore, all transitions correspond to objects that could be stacked, one on top of the other, safely and stably, or correspond to pairs of objects that were frequently placed together, side-by-side.

\subsection{Predicting a Sequence with a Modified Beam Search}
\label{section:modified_beam_search}

Regardless of how well the Markov model captures the humans' packing strategies, it is of no use without an adequate sampling mechanism.
When sampling from large stochastic models, beam search~\cite{BeamSearch} generally provides better sampling than naive approaches such as greedy search. However, we introduce modifications to adjust it to models such as ours. 

Firstly, recall that in each scene the set of objects that need to be packed is a subset of all objects, and consequently some objects in the Markov chain may be unavailable.
In this case, standard beam search cannot be applied since the search must be restricted to only the objects that remain unpacked.
This not only restricts what child nodes can be considered during the search, but if, for a set of objects that were not yet packed, a node has no possible child nodes, it may still have a valid transition to a grandchild node through an unavailable child.
By grandchild node we denote all nodes reachable from the current node with two or more transitions.
This can make the search process very slow because for each node the algorithm would need to search all of its child and grandchild nodes to find valid transitions.
The key observation of our solution is that the Markov chain is static - it never changes.
Because of this we can build and store a table of all possible transitions for each object before executing the algorithm.

\SetKwComment{Comment}{/* }{ */}
\begin{algorithm}[h]
    \label{alg:res:modified_beam_search}
% \setstretch{1.2}
    % \KwData{$n \geq 0$}
    % \KwResult{$y = x^n$}
    $beam\_width \gets 5$\;
    $transitions \gets$ load level representation of Markov chain\;
    $unpacked \gets$\: list of unpacked objects $O$ \Comment*[r]{Obtained from the VR environment} 
    $open\_list \gets$\: initial beam with $<\!start\!>$ $S$ state \Comment*[r]{Contains active search beams} 
    \While{\textnormal{not exhausted all transitions for all beams in} $open\_list$}{
        $cur\_trnst\_level \gets 1$ \Comment*[r]{Current Transition level to search in} 
        \While{\textnormal{at least one} $beam$ \textnormal{in} $open\_list$ \textnormal{has transitions for} $cur\_trnst\_level$}{
            \ForEach{$beam$ \textnormal{in} $open\_list$}{
                \If{$trnsts$ \textnormal{has valid child states for this} $beam$ \textnormal{and} $cur\_trnst\_level$}{
                    $open\_list \mathrel{{+}{=}} Gen(beam,\:trnsts[beam][cur\_trnst\_level])$\Comment*[r]{Generate new beams} 
                    remove $beam$ from $open\_list$\;
                }
            }
            \uIf{\textnormal{at least one} $beam$ \textnormal{was extended}}{
                sort $open\_list$ according to $beam$ probability\;
                keep top $beam\_width$ beams and delete the others\;
                break\;
            }
            \Else{
                $cur\_trnst\_level$ ++\;
            }
        }
    }
    \Return most likely $beam$ in $open\_list$ 
    \Comment*[r]{All valid transitions have been exhausted} 
    
    \caption{Modified beam search algorithm used to sample a sequence prediction from our model.}
\end{algorithm}

For each state in $S$~(the set of all states), we store all possible transitions in levels: 
the \textbf{first level} contains the transitions to direct child nodes -- those that require only one transition; the \textbf{second level} contains the transitions to the grandchild nodes that require two transitions, and so forth.
The original Markov chain contains only the child nodes for each state -- the minimum amount of information needed to store the chain -- while this new representation contains all transitions for all states.
This increases the algorithm's memory complexity but makes it faster, as the search has instant access to all valid transitions.

Secondly, during the search process we consider that a sequence is complete when it reaches a leaf node -- a node that has no more valid transitions.
But longer sequences will generally have increasingly smaller probabilities (given the probabilities' multiplication). 
%If at the end of an iteration, the algorithm needs to select between two sequences - one is shorter and complete, the other is longer and still incomplete - which one should be discarded?
If the search considers only the likelihood of a sequence then it will inevitably favor shorter sequences that have higher probabilities.
This effect, in conjunction with the first change we made, would encourage the algorithm to find sequences that reach a leaf node with the smallest length possible. 
Thus, we introduce a preference for longer sequences to the detriment of complete sequences.
In practice this is accomplished with two additions:
\begin{enumerate}
    \item During the search process, choose sequences that can still expand to more nodes to the detriment of sequences that have already reached a leaf node.
    \item Enforce that transitions to a new node are as small as possible.
    In other words, start by searching for a valid state in the first level of our new representation.
    If none are found, proceed to the second level, and so forth.
\end{enumerate}
The final search algorithm is presented in Algorithm \ref{alg:res:modified_beam_search}.

\section{Results}

Now that an appropriate sampling mechanism has been implemented, this section focuses on describing how to test the sampled sequences.

\subsection{Testing the Model}
The main challenge lies in the fact that, for a given set of objects, there are numerous valid packing sequences.
Depending on how the objects are placed inside the box, two distinct sequences can lead to an equally efficient packing.
This dependency on object placement makes evaluating a sequence prediction very difficult without actually placing the objects inside the box, and currently the methods to predict an object placement pose in a packing task are limited. 
As such, drawing some inspiration from the Turing test in which a human evaluator must distinguish between human or computer generated text responses, we develop a new VR environment to test the sequence prediction model.
This new environment is visually identical to the environment used to collect BoxED but the participant no longer chooses the order in which the objects are packed.
Instead, the sequence is indicated by highlighting what object should be placed next.
Some of the sequences are a replay of the real sequences from BoxED, whereas others are generated by the computer.
At the end of each scene, the participant is asked if the sequence was a real sequence executed by a human or a computer generated sequence.
%The participants answer ``human generated" if they felt that the sequence was logical and human-like, or ``computer generated" otherwise.

There are three types of computer generated sequences:
\begin{enumerate}
    \item Random: given the subset of objects on the table, a packing sequence is randomly generated.
    \item Beam-N: sample a sequence prediction using the modified beam search algorithm.
    In this case, no limit is imposed on the length of a sequence prediction.
    \item Beam-3: the same as Beam-N except a maximum sequence length of 3 is imposed.
    This method was introduced because we noticed that Beam-N produces long sequences that often pack fragile objects too soon in the packing process.    
\end{enumerate}

Each of the four types of sequence has its purpose.
The real sequences verify that the participants can detect sequences executed by other humans.
This tests whether the personal packing preferences are sufficiently strong to influence the final evaluation, or if there are general packing strategies that most humans adhere to.
The random sequences test if participants distinguish between logical and random sequences. These two types of sequences will determine if humans are good evaluators for this task and validate the experimental design.
Finally, the results from the Beam-N and Beam-3 sequences will indicate if our model was able to ``trick" participants into believing that the sequence was generated by another human.
%Note that in these two cases, a new sequence is sampled from the model whenever the participant packs all the objects in the previous sequence or when the participant places an object at the top of a stack of objects, near the top of the box.
%The intuition is that the next object will be placed next to this stack~(and not on top of it), so a new sequence should be sampled from the top of the Markov chain.

\subsection{Experimental Results}

In total this second experiment involved 29 participants for a total of 96 evaluations, shown in Table~\ref{table:seq_pred_results}.
Since Beam-N was replaced by Beam-3 during the experiment, both have fewer samples than the other two sequence types.

\begin{table}[h]
\caption{Results from sequence prediction experiment}
\centering
\resizebox{0.8\columnwidth}{!}{%
\centering
\begin{tabular}{c|cc|c}
\cline{2-3}
                                    & \multicolumn{2}{c|}{Participants evaluations}                                                                     &                                                                                  \\ \hline
\multicolumn{1}{|c||}{Sequence Type} & \multicolumn{1}{c|}{\textit{\begin{tabular}[c]{@{}c@{}}Human\\ generated\end{tabular}}} & \textit{\begin{tabular}[c]{@{}c@{}}Computer \\ generated\end{tabular}} & \multicolumn{1}{||c|}{\begin{tabular}[c]{@{}c@{}}Number of\\ samples\end{tabular}} \\ \hline
\multicolumn{1}{|c||}{Random}        & \multicolumn{1}{c|}{7\%}                 & \textbf{93\%}                                                          & \multicolumn{1}{||c|}{30}                                                          \\ \hline
\multicolumn{1}{|c||}{Real}          & \multicolumn{1}{c|}{\textbf{79\%}}       & 21\%                                                                   & \multicolumn{1}{||c|}{33}                                                          \\ \hline
\multicolumn{1}{|c||}{Beam-N}        & \multicolumn{1}{c|}{50\%}                & 50\%                                                                   & \multicolumn{1}{||c|}{16}                                                          \\ \hline
\multicolumn{1}{|c||}{Beam-3}        & \multicolumn{1}{c|}{\textbf{88\%}}       & 12\%                                                                   & \multicolumn{1}{||c|}{17}                                                          \\ \hline
\end{tabular}
}
\label{table:seq_pred_results}
\end{table}

Regarding the real sequences, the result is positive with 79\% of the evaluations correctly identifying these as \textit{human generated}, and thus we can infer that participants are generally capable of identifying a sequence produced by another human.
The percentage of incorrect answers could be attributed to individual preference or related to the fact that the participant
may place a few of the initial objects in positions that become
an obstacle or that are incompatible with the rest of the
sequence. Likewise, the vast majority correctly identified that the randomly generated sequences were illogical, proving that there are significant differences between real and random sequences.
Along with the previous result, this validates the experimental design - humans are good evaluators of this task and can recognize packing strategies in sequences produced by other humans.
This implies that there are, in fact, general rules or strategies that most humans apply to this task.

Unexpectedly, 50\% of participants correctly detected that the sequences generated by our model with Beam-N sampling were computer generated.
Such a high percentage clearly indicates that sequences sampled with Beam-N are not similar to the real sequences.
This result led to the development and introduction of the variation Beam-3. This modification clearly had a positive influence on the performance of our model.
The success rate at ``tricking" participants increased from 50\% with Beam-N to a surprising 88\% with Beam-3, even higher than the percentage of real sequences that were classified as \textit{human generated}.
One possible explanation for this result is that our model learned more general principles for the task that the majority of humans adhere to, instead of learning specific principles that only a few humans consider, thus pleasing more participants.

To validate these intuitive conclusions we conduct Boschloo's exact test~\cite{boschloo} (instead of the standard two proportions Z-test due to limited sample size) to test whether the proportion of individuals that was ``tricked" by the Beam-3 algorithm, $p_1$, is greater than the proportion of individuals ``tricked" by Beam-N, $p_2$, at a significance level of $\alpha=0.05$.
As such, our null and one-tailed alternative hypotheses are $ H_0: p_1 = p_2 $ and $ H_1: p_1 > p_2 $. 
From this test we obtain a p-value of 0.0099, and hence, at this significance level, we reject the null hypothesis and prove the difference between these two proportions.
In other words, Beam-3 is an improvement over Beam-N.
However, this only guarantees that there is an effect, and \emph{not} that it is significant.
The effect size measures importance or significance on a normalized scale, regardless of the quantities in the study.
There are multiple measures of this effect size, but we consider Cohen's H \cite{cohen1988}, as is standard in statistical analysis.
Our results produce an $h$ value of approximately 0.86, which indicates a large effect according to Cohen's interpretation \cite{cohen1988}.
Finally, these results are validated with a post-hoc power analysis, reporting a power of 79\%, which is a satisfactory power value according to~\cite{cohen1988}.

%%%%%%%%%%%%%%%%%%%%%%%%%%%%%%%%%%%%%%%%%%%%%%%%%%%%%%%%%%%%%%%%%%%%%%%%%%%%%%%%
\section{Conclusions}
\label{conclusions}

%\textcolor{red}{Improve the conclusion: overall work; major conclusion; future work; take home message;}

We introduced a new method for learning the packing sequence of irregular objects from human demonstrations.  
Our model's solutions have been consistently perceived as exhibiting human-like characteristics by participants. 
Through the direct acquisition of implicit task constraints from expert demonstrations, our work demonstrates that tapping into human knowledge not only sidesteps the need for labor-intensive and data-heavy approaches but also produces human-like behaviors.

Besides tackling this largely unsolved problem, packing a variety of groceries in a proper manner (to prevent breaking or squishing some items), we created the BoxED platform, a VR environment capable of recreating fast packing scenarios for reliable and ease-of-use data acquisition of human demonstrations. In addition to the BoxED platform, we provide one of the first datasets of packing sequence of irregular objects. By creating the first dataset of demonstrations of human experts packing groceries we hope to stimulate more research into automating this task and into the broader area of collaborative robots.

Future work will encompass the development of a supervised learning approach for predicting the placement pose of objects in the bin packing task, along with the introduction of a level of abstraction to obviate the need for explicit knowledge of each object's physical properties.

%Using a portion of our new dataset our model surpassed human performance at predicting object packing sequences that are classified by experts as human-like. 

%% COMMENTS FROM JSV
%refs ? sounds a bit vague...
% falta fazer o recap dos achievements e contributions

%%%%%%%%%%%%%%%%%%%%%%%%%%%%%%%%%%%%%%%%%%%%%%%%%%%%%%%%%%%%%%%%%%%%%%%%%%%%%%%%

\bibliographystyle{IEEEtran}
\bibliography{IEEEabrv, bibliography}

\end{document}